\documentclass{article}
\usepackage{spconf,amsmath,graphicx}
\usepackage{comment}
\usepackage{courier}
\usepackage{amsmath}
\usepackage[mathscr]{euscript}
\usepackage{subcaption}
\usepackage{hyperref}

\usepackage{babel}
\usepackage{verbatim}
\usepackage{caption}
\usepackage{subcaption}
\usepackage{multicol}
\usepackage{graphicx}

\usepackage[dvipsnames]{xcolor}
\usepackage{booktabs} % To
\usepackage{multirow}

\usepackage[symbol]{footmisc}

\renewcommand{\paragraph}[1]{\noindent\textbf{#1}\quad}

\def\L{{\cal L}}

\title{Language Model is All You Need: Natural Language Understanding As Question Answering}
\name{Mahdi Namazifar, Alexandros Papangelis, Gokhan Tur, Dilek Hakkani-T\"ur}
\address{Amazon Alexa AI}
\begin{document}
\ninept
\maketitle
\begin{abstract}
Different flavors of transfer learning have shown tremendous impact in advancing research and applications of machine learning. In this work we study the use of a specific family of transfer learning, where the target domain is mapped to the source domain.  Specifically we map Natural Language Understanding (NLU) problems to Question Answering (QA) problems and we show that in low data regimes this approach offers significant improvements compared to other approaches to NLU. Moreover we show that these gains could be increased through sequential transfer learning across NLU problems from different domains. We show that our approach could reduce the amount of required data for the same performance by up to a factor of 10.
\end{abstract}
\begin{keywords}
Transfer Learning, Question Answering, Natural Language Understanding
\end{keywords}
%

%%%%%%%%%%%%%%%%%%%%%%%%%
%%%%%%%%%%%%%%%%%%%%%%%%%
%%%%%%%%%%%%%%%%%%%%%%%%%
\vspace{-0mm}
\section{Introduction}
\label{sec:intro}

\begin{figure*}[t]
  \centering
  \begin{tabular}{c|c|c}
    \centering
      \begin{subfigure}{0.29\textwidth}
        \includegraphics[width=\textwidth]{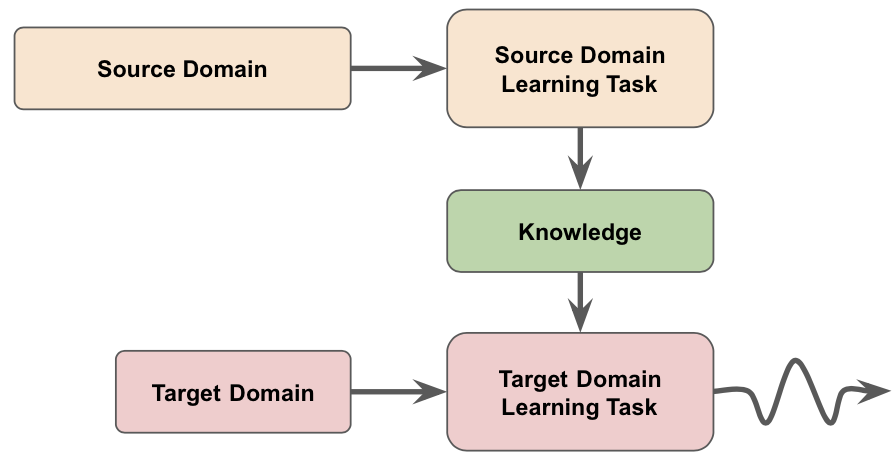}
          \caption{\footnotesize Transfer Learning from source domain to target domain}
          \label{fig:transfer}
      \end{subfigure} &
      \centering
      \begin{subfigure}{0.29\textwidth}
        \includegraphics[width=\textwidth]{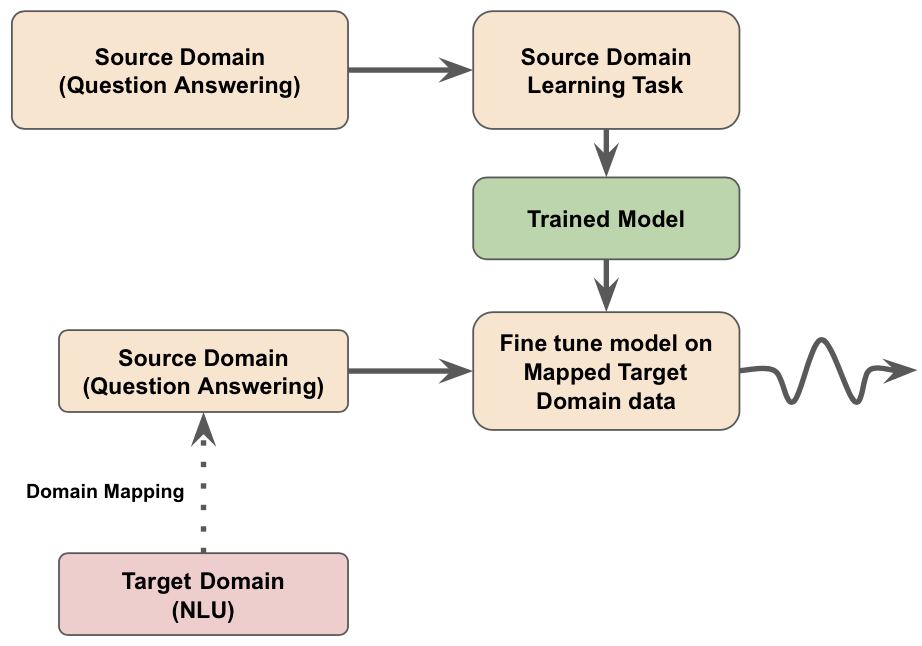}
          \caption{\footnotesize Transfer learning through mapping a target domain to source domain. In this work we map NLU to QA tasks.}
          \label{fig:target_map}
      \end{subfigure}&
      \centering
      \begin{subfigure}{0.36\textwidth}
        \includegraphics[width=\textwidth]{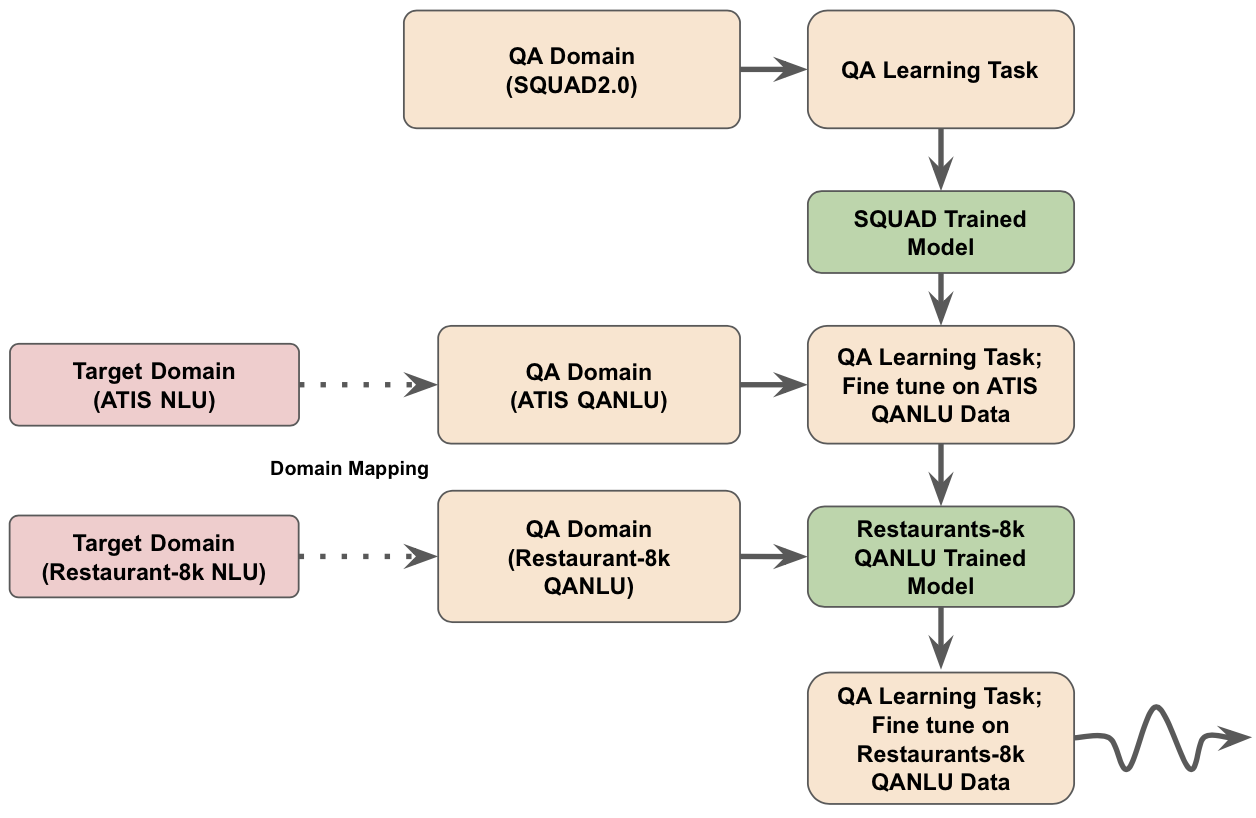}
          \caption{\footnotesize Sequential transfer learning for QANLU.}
          \label{fig:atis_to_r8k}
      \end{subfigure}
    %   & 
    %   \centering
    %   \begin{subfigure}{0.31\textwidth}
    %     \includegraphics[width=\textwidth]{data_aug.png}
    %       \caption{\footnotesize Data augmentation for source domain as a consequence of mapping target domain to source domain}
    %       \label{fig:data_aug}
    %   \end{subfigure}
   \end{tabular}
   \caption{\vspace{-6mm}}
   \label{fig:transfer_learning}

\end{figure*}

Transferring the knowledge that machine learning models learn from a source domain to a target domain, which is known as transfer learning (Figure \ref{fig:transfer}) \cite{10.1007/978-3-030-01424-7_27, zhuang2020comprehensive}, has shown tremendous success in Natural Language Processing (NLP) \cite{Alyafeai2020ASO, pennington2014glove,NIPS2013_5021, devlin-etal-2019-bert, Raffel2019ExploringTL}. 

One of the most prominent advantages of transfer learning is manifested in low data regimes. 
As the models become increasingly complex, in most cases this complexity comes with requirements for larger training data which makes transferring the learning from a high data domain to a low data domain very impactful. In this work we focus on the type of transfer learning in which the target domain is first mapped to the source domain. Next a model is trained on the source domain. Then the transfer of knowledge is done through fine-tuning of this model on the mapping of the target domain (to the source domain), as shown in Figure \ref{fig:target_map}.  As an example of this transfer learning paradigm in NLP, decaNLP \cite{DBLP:journals/corr/abs-1806-08730} could be mentioned where 10 NLP tasks are mapped to the Question Answering (QA) problem, in which given a context the model should find the answer to a question.

% \begin{figure*}[t]
%   \centering
%   \begin{tabular}{|c|c|c|}
%       \subcaptionbox{Transfer Learning}[.3\linewidth][c]{%
%         \includegraphics[width=1\linewidth]{transfer_learning.png}}\quad
%         \label{fig:transfer}
%       & \subcaptionbox{Transfer learning through mapping target domain to  source domain. In this work we map NLU to QA tasks.}[.3\linewidth][c]{%
%         \includegraphics[width=1\linewidth]{target_map.png}}\quad
%       & \subcaptionbox{Data augmentation for source domain as a consequence of mapping target domain to source domain}[.3\linewidth][c]{%
%         \includegraphics[width=1\linewidth]{data_aug.png}}
%   \end{tabular}
%   \caption{Transfer Learning}
%   \label{fig:transfer_learning}

% \end{figure*}

In this work, we map Natural Language Understanding (NLU) problems to the QA problem. Here NLU refers to determining the intent and value of slots in an utterance \cite{DBLP:journals/corr/abs-1902-10909}. For instance in ``show cheap Italian restaurants'' intent could be \textit{inform} and the value for slot \textit{cuisine} is ``Italian'' and for slot \textit{price range} is ``cheap''. More specifically in our approach to which we refer as QANLU, we build slot and intent detection questions and answers based the the NLU annotated data. QA models are first trained on QA corpora and then fine-tuned on questions and answers created from NLU annotated data. In this approach transfer learning happens through transferring knowledge of finding the answer to a question given a context, that is acquired by the model during the training of the QA model, to finding the value of an intent or a slot in text input. Through our computational results we show that QANLU in low data regimes and few-shot settings significantly outperforms the sentence classification and token tagging approaches for intent and slot detection tasks, as well as the newly introduced ``IC/SF few-shot'' approach \cite{krone-etal-2020-learning} for NLU. We also show that QANLU sets a new state of the art performance on slot detection on the Restaurants-8k dataset \cite{coope2020spanconvert}. Furthermore, we show that augmenting the QA corpora with questions and answers created based on NLU annotated data improves the performance of QA models. Throughout this work we use span selection based QA models built on top of transformer-based language models \cite{devlin-etal-2019-bert}. That being said, our approach is quite generic and could be extended to any type of QA system.
\begin{comment}
\begin{itemize}
    \item Importance of transfer learning; specially in low data regimes and few-shot learning
    \item Question Answering is a generic task to which many tasks could get mapped to.
    \item Prior works on mapping different NLP tasks to QA
    \item QA as an oracle
    \item QA for NLU challenges; both precision and recall should be high
    \item Our approach
    
\end{itemize}
\end{comment}

%%%%%%%%%%%%%%%%%%%%%%%%%
%%%%%%%%%%%%%%%%%%%%%%%%%
%%%%%%%%%%%%%%%%%%%%%%%%%
\section{Related Works}
Framing NLP tasks as QA has been studied in the past. For instance \cite{DBLP:journals/corr/abs-1806-08730} maps 10 NLP tasks (excluding intent and slot detection) into QA and trains a single model for all of them. However, this work does not explore the task of intent and slot classification. In a similar line of reasoning, \cite{gao2019dialog} poses the Dialogue State Tracking (DST) task as machine reading comprehension (MRC), formulated as QA. \cite{gao2020machine} builds on that work achieving competitive DST results with full data and in few-shot settings. \cite{zhou2019multi} also explores DST as QA, using candidate values for each slot in the question (similar to the Multiple-Choice setting of \cite{gao2020machine})  achieving slightly better results than \cite{gao2020machine}. We propose a method that is conceptually similar but focuses on low-resource applications and does not require designing and training of a new model architecture or extensive data pre-processing, achieving strong results in slot and intent detection with an order of magnitude less data. Here we do not discuss all intent or slot detection methods. However, some notable few-shot NLU works include \cite{bapna2017towards,bhathiya2020meta,shah2019robust,coope2020spanconvert, bapna2017towards}, and we compare against their results when appropriate. Other interesting approaches that do not require training include priming pre-trained language models, e.g. \cite{madotto2020language}.

%%%%%%%%%%%%%%%%%%%%%%%%%
%%%%%%%%%%%%%%%%%%%%%%%%%
%%%%%%%%%%%%%%%%%%%%%%%%%
\section{Question Answering for Natural Language Understanding (QANLU)}
\label{sec:map}

\subsection{Slot Detection}
\label{sec:data_prep}
Consider a set of text records $T = \{t_1, t_2, ..., t_n\}$ in which each record is annotated for the set of slots $S = s_1, s_2, ..., s_m$. Also for each slot $s_j$ consider a set of questions $Q_{s_j}=\{q_{s_j1}, q_{s_j2}, ..., q_{s_jk_j}\}$ that could be asked about $s_j$ given any text record $t_i$. The following is an example of such a setting:

\begin{equation*}
\footnotesize
\begin{aligned}
&S:\{\mbox{\texttt{\scriptsize{cuisine}}}, \mbox{\texttt{\scriptsize{price range}}}, \mbox{\texttt{\scriptsize{area}}}\}, t_i:  \mbox{\textit{``Show cheap Italian restaurants''}} \\
&\hspace{1mm} \mbox{\texttt{\footnotesize{cuisine}}: ``Italian''}, \hspace{1mm} \mbox{\texttt{\footnotesize{price range}}: ``cheap''}, \hspace{1mm}\mbox{\texttt{\footnotesize{area}}: ``''} \\
&Q:\{Q_{\mbox{\texttt{\scriptsize{cuisine}}}}, Q_{\mbox{\texttt{\scriptsize{price range}}}}, Q_{\mbox{\texttt{\scriptsize{area}}}}\}
\end{aligned}
\end{equation*}
where

\begin{equation*}
\footnotesize
\begin{aligned}
& Q_{\mbox{\texttt{\scriptsize{cuisine}}}}:  \hspace{0mm}\{\mbox{``what cuisine was mentioned?''},  \\
& \hspace{17mm}\mbox{``what type of food was specified?''}\} \\
& Q_{\mbox{\texttt{\scriptsize{price range}}}}: \{\mbox{``what price range?''}\}  \\
& Q_{\mbox{\texttt{\scriptsize{area}}}}: \hspace{0mm}\{\mbox{``what part of town was mentioned?''}, \mbox{``what area?''}\}
\end{aligned}
\end{equation*}

Given $T$, $S$, and $Q$ it is straightforward to create the set of all the possible questions and their corresponding answers for each $t_i$ as the context for the questions:
%\begin{equation*}
\vspace{-2mm}
\begin{align*}
\footnotesize
\mbox{\bf{Context:}}\hspace{4mm}\mbox{\textit{``Show cheap Italian restaurants''}} \\
\mbox{what cuisine was mentioned?}&       \hspace{4mm}\mbox{``Italian''} \\
\mbox{what type of food was specified?}& \hspace{4mm} \mbox{``Italian''}  \\
\mbox{what price range?}& \hspace{4mm} \mbox{``cheap''}  \\
\mbox{what part of town was mentioned?}& \hspace{4mm} \mbox{``''}  \\
\mbox{what area?}& \hspace{4mm} \mbox{``''}
\end{align*}
We experiment with different ways of creating the set $Q$. This set could be handcrafted, i.e. for each slot we create a set of questions separately, or created using templates such as ``what \underline{\hspace{10mm}} was mentioned?'' where we the blank is filled with either the slot name or a short description of the slot, if available.
\vspace{-3mm}
\subsection{Intent Detection}

For intent detection we add ``yes. no.'' at the beginning of the context and for each intent we create a question like ``is the intent asking about \underline{\hspace{10mm}}?'' where the blank is filled with the intent. The answer to these questions are ``yes'' or ``no'' from the segment that was added to the beginning of the context depending on whether the intent is in the context or not. 

\vspace{-3mm}
\subsection{Question Answering Model}
In this work we use span detection based QA models that are built on top of transformers \cite{NIPS2017_7181} as are described in \cite{devlin-etal-2019-bert}. We also use the SQuAD2.0 \cite{rajpurkar-etal-2018-know} data format for creating questions and answers, as well as the corpus for the source domain (QA). Note that in converting annotated NLU data to questions and answers in QANLU, since for each text record we ask all the questions for all the slots (whether they appear in the text or not), many of the questions are not answerable. As was discussed earlier, we use pre-trained QA models that are trained on SQuAD2.0 (the green box in Figure \ref{fig:target_map}) and fine-tune them with the questions and answers that are created from the NLU tasks. We also study how in a sequential transfer learning style we can improve the performance of NLU through QANLU (Figure \ref{fig:atis_to_r8k}).

%%%%%%%%%%%%%%%%%%%%%%%%%
%%%%%%%%%%%%%%%%%%%%%%%%%
%%%%%%%%%%%%%%%%%%%%%%%%%

\section{Computational Results}
In this section we present our computational results for QANLU. Our experiments are done on the ATIS \cite{hemphill-etal-1990-atis, 5700816} and Restaurants-8k \cite{coope2020spanconvert} datasets.  All of the experiments are implemented using Huggingface \cite{Wolf2019HuggingFacesTS}, and we also use pre-trained language models and QA models provided by Huggingface and fine-tune them for our QA data. We base our experiments mainly on pre-trained DistilBERT \cite{sanh2020distilbert} and ALBERT \cite{Lan2020ALBERTAL} models%\footnote{We will release the all the data that we have created upon publication of this work.}.
.
\subsection{ATIS}
\vspace{-2mm}
The ATIS dataset is an NLU benchmark that provides manual annotations for  utterances inquiring a flight booking system. 
%There are 84 different slots and 21 different intents in the ATIS dataset. 
Since the original ATIS dataset does not have a validation set, we use the split of the original training set into training and validation that is proposed in \cite{zhang2019joint}. 
For each slot in ATIS we create a question set and for each record in ATIS we create the set of questions and answers based on all the question sets and the slot and intent annotation of the record, according to the approach described in Section \ref{sec:map}. In the first set of experiments we study how our QANLU approach compares to the widely used joint token and sentence classification \cite{DBLP:journals/corr/abs-1902-10909} in few-shot settings using different stratification in sampling of the training records for the few-shot setting. Table \ref{tbl:fewshot} summarizes the results. In this table we report F1 scores for both slots and intent detection tasks. The reason why we use F1 scores for intent detection is that in the ATIS dataset each record could have more than one intent. Each value in Table \ref{tbl:fewshot} is an average over 5 runs with different random seeds. Each row in this table represents one sample of the ATIS training data. The  set of rows titled ``\(\mathcal{N}\) uniform samples'' are sampled uniformly with samples of sizes of 10, 20, 50, 100, 200, and 500 ATIS records. The set of rows titled ``\(\mathcal{N}\) samples per slot'' are sampled such that each sample includes at least \(\mathcal{N}\) instances for any of the slots, where \(\mathcal{N}\) is 1, 2, 5, or 10. The set of rows titled ``\(\mathcal{N}\) samples per intent'' are sampled such that each intent appear in at least \(\mathcal{N}\) instances, where \(\mathcal{N}\) is 1, 2, 5, or 10. The numbers in parenthesis in front of \(\mathcal{N}\) represent the number of ATIS records in the sample. For each ATIS record we have 179 questions and answers for intents and slots. 

In Table \ref{tbl:fewshot} we report performance of models based on both DistilBERT and ALBERT. For QANLU we fine-tune a QA model trained on SQuAD2.0 data (``distilbert--base--uncased--distilled--squad''\footnote[2]{Model acquired from \url{www.huggingface.co/models}\label{hf}}  for DistilBERT and ``twmkn9/albert--base--v2--squad2''\footref{hf} for ALBERT) on our questions and answers for ATIS samples. We Also train joint intent and token classification models for the ATIS training samples based on pre-trained DistilBERT  and ALBERT models (``distilbert--base--uncased''\footref{hf} and ``albert--base--v2''\footref{hf})\footnote[3]{We also tried these models fine-tuned on SQuAD2.0, but they didn't perform as well on the intent and token classification tasks}. We compare the results of QANLU models  with the classification based models (noted as QANLU and Cls in the table, respectively). It is clear that QANLU models outperform classification based models, often by a wide margin. For instance for the ALBERT based model, for the case where there is at least 1 sample per slot the QA based model outperforms the classification based model by 26\% (86.37 vs 68.26). It is notable that the gap between the two approaches narrows as the number of samples increases, with the exception of intent detection for the uniform sample with only 10 samples. In a closer look at this sample, the intent for all the records is the same (``atis\_flight'' which is the intent for 74\% of the ATIS training set) and that could explain why the models almost always predict the same value for the intent.

The fact that for both DistilBERT and ALBERT based models we see that the QANLU significantly outperforms the intent and slot classification models in few-shot settings indicates that the performance improvements are likely stemmed from transfer learning from reading comprehension that is learned in the QA task.

In this set of experiments we used handcrafted questions for each slot. One could argue that creating questions for slots is as difficult or perhaps more difficult as getting data annotated specifically for intents and slots. To see if we can detour the manual question creation process we also experimented with questions that were created using frames based on a brief description of each slot as well as using the tokenized slots names. These frame based questions could be easily created for free by running some simple scripts. The experimental results show no significant degradation in the performance of QANLU models trained on frame based questions. 

In another set of experiments we compare QANLU with another few-shot approach (few-shot IC/SF) proposed in \cite{krone-etal-2020-learning}.  We use the exact same split of the ATIS dataset that is created in that paper. Results are in Table \ref{tbl:ic/sf}.

\begin{table}[ht!]
%\tiny
\footnotesize
\centering

\begin{tabular}{|c|c|c|}
\hline
& Few-shot IC/SF & QANLU\\
\hline
F1 score & 43.10 & 68.69 \\

\hline
\end{tabular}
\caption{\footnotesize QANLU vs Few-shot IC/SF \cite{krone-etal-2020-learning} Slot detection F1. 43.10 is reported in Table 5 of \cite{krone-etal-2020-learning} \vspace{-3mm}}
\label{tbl:ic/sf}
\end{table}

The few-shot IC/SF results (43.10) are average of multiple runs of a BERT model first pre-trained on the training set, and then fine-tuned on a ``support'' set sampled from the test set, and then evaluated on a ``query'' set also sampled from the test set. We used the exact same training set that used in that work and trained a BERT (base size) based QANLU model on the training set. We then directly evaluated that model on the exact same test set created in \cite{krone-etal-2020-learning}, without any fine-tuning on a support set. The resulting F1 score (68.98) is 60\% higher than what is reported \cite{krone-etal-2020-learning}.
\begingroup
\setlength{\tabcolsep}{4mm} % Default value: 6pt
\renewcommand{\arraystretch}{1} % Default value: 1

\begin{table*}[ht!]
\scriptsize
%\tiny
\centering
\begin{tabular}{|c|c|c|c|c|c|c|c|c|c|}
\hline
%\vspace{-10}

& & \multicolumn{4}{c|}{\bf{Intent}} & \multicolumn{4}{c|}{\bf{Slot}} \\
\hline
&  & \multicolumn{2}{c|}{DistilBERT} &   \multicolumn{2}{c|}{ALBERT} & \multicolumn{2}{c|}{DistilBERT} &
\multicolumn{2}{c|}{ALBERT}\\
%%%%%%%%%%%%%%%

\hline
 & \( \mathcal{N} \) & QANLU &	Cls & QANLU &	Cls &	QANLU &	Cls &	QANLU &	Cls \\
\hline

\multirow{6}{*}{\( \mathcal{N} \) uniform} &10&71.80	&71.78	&72.18	&71.78	&\bf{67.23}	&61.60	&\bf{64.24}	&54.78\\
\multirow{6}{*}{samples} & 20&\bf{83.95}	&77.80	&\bf{83.28}	&75.36	&\bf{78.53}	&56.70	&\bf{74.53}	&51.67\\
&50 &\bf{86.07}	&78.93	&\bf{86.32}	&73.90	&\bf{83.84}	&76.61	&\bf{80.26}	&74.04\\
&100 &\bf{93.08}	&87.91	&\bf{92.14}	&80.20	&\bf{85.69}	&80.34	&\bf{83.13}	&77.50\\
&200 &\bf{94.30}	&90.97	&\bf{96.78}	&85.02	&\bf{91.24}	&85.32	&\bf{89.57}	&83.63\\
&500 &\bf{96.40}	&95.45	&\bf{96.77}	&90.62	&\bf{92.31}	&91.15	&\bf{91.18}	&86.69\\

\hline

\multirow{4}{*}{\( \mathcal{N} \) samples per} &1 (75)	& \bf{88.72}	&86.47	&\bf{90.91}	&84.93	&\bf{88.47}	&76.24	&\bf{86.37}	&68.26 \\
\multirow{4}{*}{slot (Total)} &2 (136)	& \bf{91.68}	&84.91	&\bf{92.11}	&82.42	&\bf{90.77}	&84.42	&\bf{90.17}	&79.49 \\
&5 (302)	& \bf{94.34}	&93.74	&\bf{95.52}	&87.47	&\bf{93.11}	&91.38	&\bf{87.82}	&86.50 \\
&10 (549)	& \bf{97.10}	&96.19	&\bf{94.21}	&92.73	&\bf{94.11}	&93.93	&\bf{92.27}	&91.68 \\

\hline

\multirow{4}{*}{\( \mathcal{N} \) samples per} &1 (17)	&\bf{40.32}	&27.91	&\bf{54.49}	&25.73	&\bf{62.57}	&55.38	&\bf{62.22}	&51.05\\
\multirow{4}{*}{intent (Total)}&2 (33)	&\bf{78.24}	&47.20	&\bf{62.22}	&23.52	&\bf{75.39}	&65.09	&\bf{74.99}	&61.01\\
&5 (81)	&\bf{86.49}	&74.08	&\bf{89.36}	&41.28	&\bf{84.40}	&80.25	&\bf{82.70}	&71.83\\
&10 (152)	&91.23	&91.16	&\bf{90.13}	&68.93	&\bf{88.37}	&83.40	&\bf{86.32}	&78.25\\

\hline
All & N/A (4478) & 98.23 & 98.37 & 97.59 & 97.90 & 95.70 & 95.80 & 94.48 & 95.37\\
\hline
\end{tabular}
\caption{\footnotesize QANLU vs. intent and token classification (Cls) \cite{DBLP:journals/corr/abs-1902-10909} for ATIS in few-shot settings. Each row is associated with a different sampling size and strategy of ATIS data. Values in bold represent statistically significant difference at p-value 0.05. Note that QANLU performs significantly better (in some cases by more the 20\%) compared to joint intent and slot classification.\vspace{-3mm}}
\label{tbl:fewshot}
\end{table*}
\endgroup

\subsection{Restaurants-8k} 
\subsubsection{QANLU for Restaurants-8k}
The Restaurants-8k dataset \cite{coope2020spanconvert} is a set of annotated utterances coming from actual conversations in the restaurant booking domain. The dataset only contains the user side utterances and slot (5 in total) annotations. The system side of the conversations are missing, but given the set of slots that are annotated at every user turn, using simple frames we can build a full context for token classification and QANLU approaches.

% for date, time, first name, last name, and number of people for a restaurant reservation. 
% The system side of the conversations are missing, but given the set of slots that are annotated at every user turn, using simple frames we can build a full context for our question answering system (QANLU). For instance for the user utterance ``I want a table for 2 at noon.'' where the annotation specifies ``people'' and ``time'' slots, we can have a frame for the combination ``people'' and time that looks like: ``system asked the user for what time and for how many people. user said I  want  a  table  for  2  at  noon''. There are 18 different slot combinations in the data for which we manually build such frames, and using these frames for each record we build a context. 

The rest of data preparation process is identical to what we described in Section \ref{sec:data_prep}.  We take both uniform and stratified samples of the training data to create few-shot settings for training QANLU models, and compare the results with token classification models. The QANLU model is again a QA model trained on SQuAD2.0 (``distilbert--base--uncased--distilled--squad''\footref{hf}) that we fine-tune on the sampled training sets. The token classification model is built on top of ``distilbert--base--uncased''\footref{hf}. The results are captured in the curves ``QANLU (SQ\(\rightarrow\)R8k)'' (SQ stands for SQuAD2.0 and R8k stands for Restaurants-8k) and ``Cls'' (stands for token classification and similar to the ATIS case is based on \cite{DBLP:journals/corr/abs-1902-10909} without the sentence classification head) in Figure \ref{fig:seq_trans}. We discuss the results in the next subsection.

\subsubsection{Sequential Transfer Learning from ATIS to Restaurants-8k} 
In another set of experiments we study whether QANLU would enable transfer learning from one NLU domain to another. This is referred to as sequential transfer learning in the literature. For this purpose we fine-tune a QANLU model that was trained on the entire ATIS training set, on samples of Restaurants-8k dataset. 
% Figure \ref{fig:atis_to_r8k} depicts the transfer learning process for this setting. 
We compare the performance of the resulting model with QANLU first trained on SQuAD2.0 and then fine-tuned on Restaurants-8k samples, as well as the token classification model.  

\subsubsection{Restaurants-8k Results}
\begin{figure}[t!]
    \centering
    \includegraphics[width=0.48\textwidth]{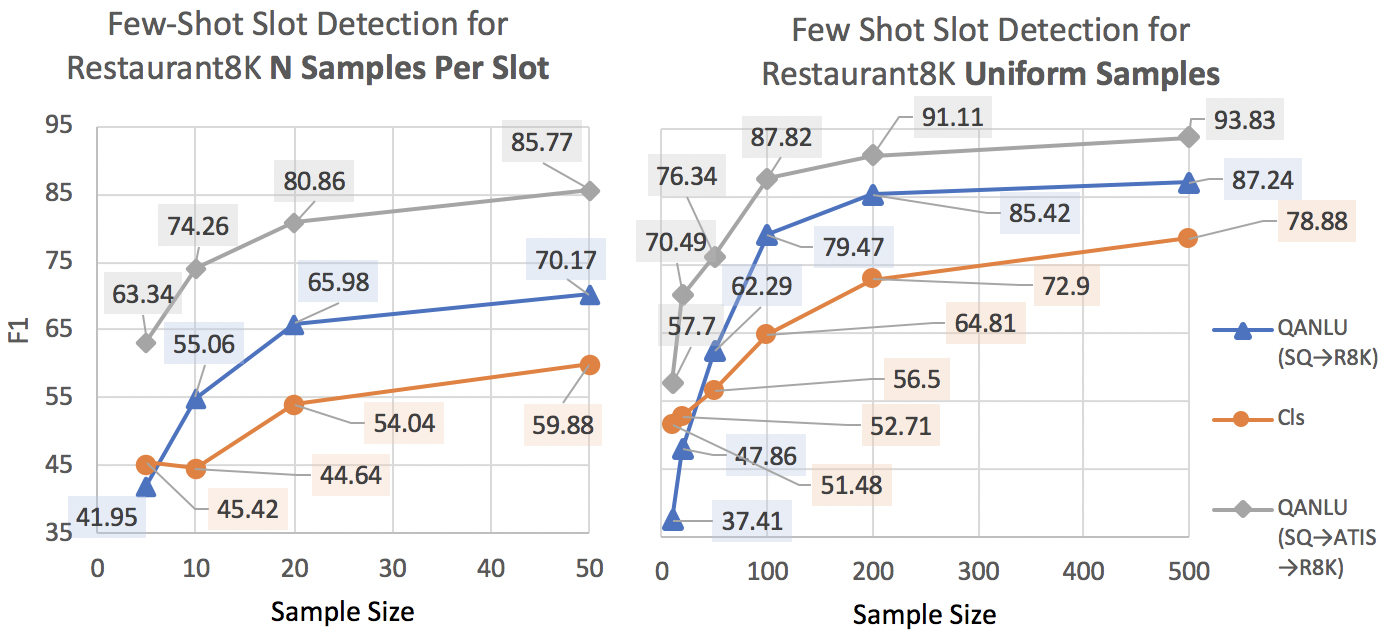}
    \caption{\footnotesize Slot detection with QANLU vs token classification. SQ\(\rightarrow\) R8k indicates QANLU first trained on SQuAD2.0 and the fine-tuned on samples of Restaurants-8k. SQ\(\rightarrow\)ATIS\(\rightarrow\) R8k is QANLU first trained on SQuAD2.0, then fine-tuned on entire ATIS, and then fine-tuned on samples of Restaurants-8k (sequential transfer learning). Cls is for the token classification approach. Numbers associated with each point are F1 scores.\vspace{-4mm}}
    \label{fig:seq_trans}
\end{figure}

In Figure \ref{fig:seq_trans} the curve QANLU (SQ \(\rightarrow\) ATIS \(\rightarrow\) r8k) is the squential transfer learning model based on ``distilbert--base--uncased--distilled--squad''\footref{hf} model (DistilBERT base model trained on SQuAD2.0). From the figure we can see that except for 10 and 20 uniform samples, for all the samples fine-tuning of SQuAD2.0 QA models on Restaurants-8k results in significantly higher F1 scores compared to the token classification approach. For uniform samples of size 10 and 20 the QANLU model (trained on SQuAD2.0 and fine-tuned on Restaurants-8k samples) performs poorly. Our intuition on the reason behind this poor performance is the small number of questions and answers for these samples (15 per record), and most likely it is not sufficient for the model to learn how to handle NLU style questions. On the other hand for the sequential transfer learning QANLU model (SQ\(\rightarrow\)ATIS\(\rightarrow\)R8k column of Figure \ref{fig:seq_trans}) we see that the model outperforms both the token classification model and the QANLU model trained on SQuAD2.0 and fine-tuned on Restaurants-8k samples by a wide margin (in some cases by over 50\%). These numbers are also shown in Figure \ref{fig:seq_trans}. This suggests that perhaps using QA as the canonical problem where NLU problems from different domains could be mapped to, could facilitate transfer learning across these NLU problems specially in few-shot settings. Also note that when the entire data is used for training the performance difference vanishes (96.98 for SQ\(\rightarrow\) R8k, 96.43 for SQ\(\rightarrow\) ATIS \(\rightarrow\)R8k, and 95.94 for Cls), which suggests that the QANLU approach is as strong as the state of the art outside of few-shot settings.

\begin{figure}[t!]
    \centering
    \includegraphics[width=0.45\textwidth]{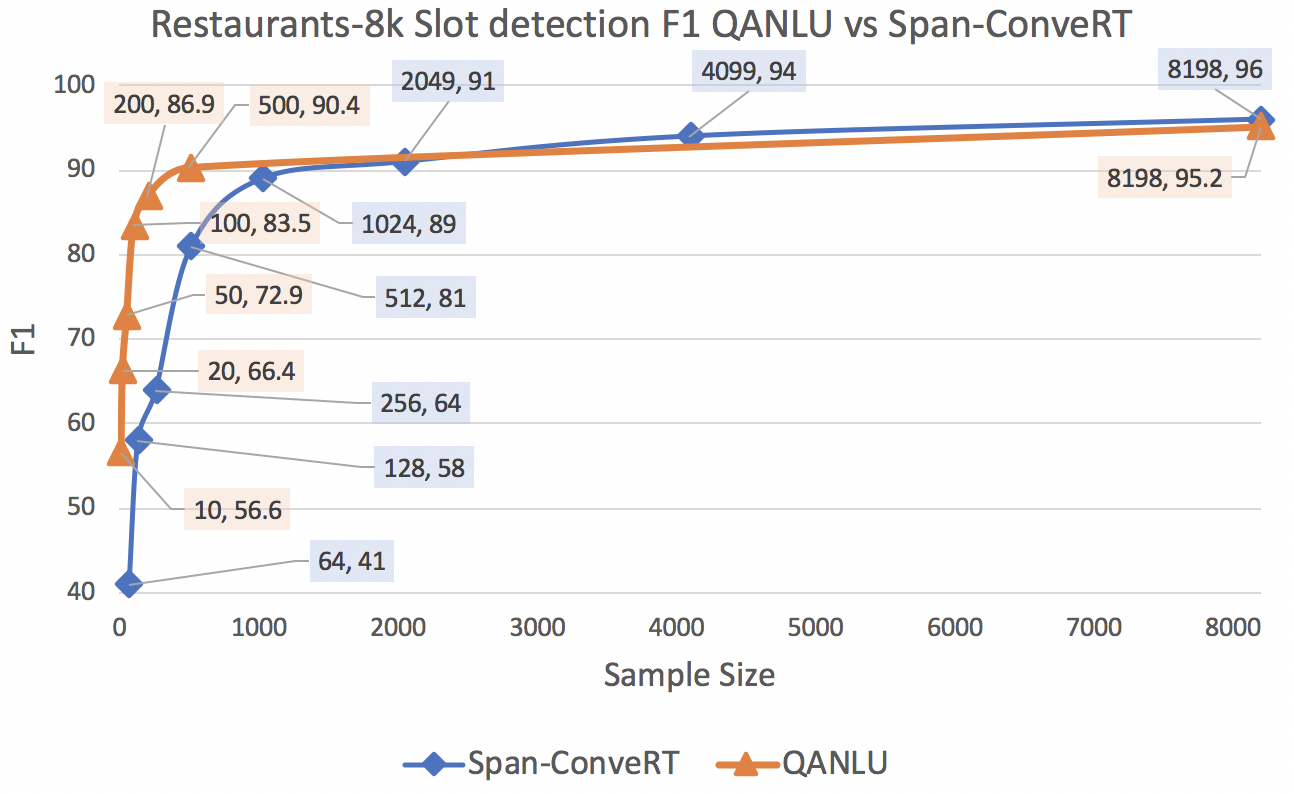}
    \caption{\footnotesize QANLU compared to Span-ConveRT \cite{coope2020spanconvert} in few-shot settings. The numbers associated with each point are the sample size and F1, respectively. \vspace{-7mm}}
    \label{fig:span-convert}
\end{figure}

Also Figure \ref{fig:span-convert} shows a comparison between QANLU and Span-ConveRT \cite{coope2020spanconvert} in few-shot settings. The few-shot F1 scores of Span-ConveRT on Restaurants-8k are borrowed from Table 3 of \cite{coope2020spanconvert}. In these experiment in order to match the settings of Span-ConverRT we do not create the previous turn for the context, hence the difference between QANLU numbers in Figure \ref{fig:span-convert} compared to Figure \ref{fig:seq_trans}. From this figure it is notable that with 20 data points QANLU reaches the higher performance than Span-ConveRT achieves with 256 data points, which translates to a 10x reduction in the amount of data needed. Also with the entire training set QANLU performs within less than 1\% of the state-of-the-art.

% \begin{table}[ht]
% \scriptsize
% %\tiny
% \centering
% \begin{tabular}{|c|c|c|c|c|}
% \hline
% & & QANLU &  & QANLU\\
% & \( \mathcal{N} \)& (SQ\(\rightarrow\) R8k) & Cls & (SQ\(\rightarrow\) ATIS \(\rightarrow\)R8k)\\
% \hline

% \multirow{6}{*}{\( \mathcal{N} \) uniform} & 10 & 37.41 & 51.48 & 57.70 \\
% \multirow{6}{*}{samples} & 20 & 47.86 &  52.71 & 70.49 \\
% & 50  & 62.29 & 56.50 & 76.34\\
% & 100 & 79.47 & 64.81 & 87.82\\
% & 200 & 85.42 & 72.90 & 91.11\\
% & 500 & 87.24 & 78.88 & 93.83\\

% \hline

% \multirow{4}{*}{\( \mathcal{N} \) samples per} & 1 (5) & 41.95 & 45.42 & 63.34	\\
% \multirow{4}{*}{slot (Total)} & 2 (10)	 &  55.06 & 44.64 & 74.26\\
% & 5 (25)   & 65.98 & 54.04 & 80.86\\
% & 10 (50)  & 70.17 & 59.88 & 85.77 \\

% \hline
% All &N/A (8198) & 96.98 & 95.94 & 96.43 \\
% \hline
% \end{tabular}
% \caption{\footnotesize QANLU vs token classification for Restaurants-8k in few-shot settings for different training set sampling strategies. Values are F1-scores and SQ stansd for SQuAD2.0 and R8K stands for Restaurants-8k.}
% \label{tbl:r8k}
% \end{table}

\vspace{-1mm}

% \begin{figure*}
% \begin{multicols}{2}
%     \includegraphics[width=0.30\textwidth]{lc_uniform.png}\par 
%     \includegraphics[width=0.30\textwidth]{lc_strat.png}\par 
%     \end{multicols}
% \caption{caption here}
% \end{figure*}
%%%%%%%%%%%%%%%%%%%%%%%%%
%%%%%%%%%%%%%%%%%%%%%%%%%
%%%%%%%%%%%%%%%%%%%%%%%%%

\section{Discussion}

The customary feeding token embeddings of a sentence into a network and mapping the output of the network for each token onto a certain number of classes for NLU seems somewhat far from our intuition on how humans understand natural language. The main research question that we try to answer is whether all NLP problems can be efficiently and effectively mapped to one canonical problem. If the answer is yes, could that canonical problem be QA? In this work we scratch the surface on these questions, in that we showcase the strength of transfer learning that happens in this paradigm in learning from few examples for intent and slot detection. But our experiments were limited to span detection QA problem and SQuAD2.0 QA data. Future works will include going beyond this configuration and also expanding across different NLP problems. Measuring how much transfer of knowledge could be achieved across different NLP tasks would be interesting to know. Another future direction could be studying how the questions for QANLU could be generated automatically based on the context.

One interesting side product of QANLU is that the questions and answers created for NLU tasks could augment the questions and answers of the QA task (SQuAD2.0 in this work) in order to improve the QA model performance. To study this idea we used the exact training script that Huggingface provides for training QA models on the SQuAD2.0 and also the SQuAD2.0 augmented with questions and answers that we created for ATIS QANLU. The training scripts specify 2 training epochs. It could be argued that this comparison would not be fair since 2 passes over the augmented data means a lot more optimization steps since there are many more questions and answers in the augmented data. To account for this we also run the training on the original SQuAD2.0 data for the same number of optimization steps as it takes to run 2 epochs on the augmented data (9000 steps). The results (QA F1 on the validation set) are shown in Table \ref{tbl:data_aug}. As the numbers show training the same models on the augmented data significantly improves the performance of the final QA model on the Development set of SQuAD2.0. We believe this result could be an indication that we can not only transfer from QA to other NLU tasks, we can also improve QA through data augmentation by mapping NLU problems to QA.

\begin{table}[ht!]
%\tiny
\scriptsize
\centering

\begin{tabular}{|c|c|c|c|}
\hline
& SQuAD2.0 & SQuAD2.0 + ATIS & SQuAD2.0\\
& (2 epochs) & (2 epochs = 9k steps & (9k steps)\\
% &  & 9,000 steps) & \\
\hline
``bert-base-cased'' & 70.07 & 74.29 & 65.42 \\
``distilbert-base-uncased''	& 55.58	& 60.26	& 57.03\\
``albert-base-v2''	& 78.05	& 79.26	& 76.44\\

\hline
\end{tabular}
\caption{\footnotesize F1 scores of QA models on original SQuAD2.0 and the augmented SQuAD2.0 with ATIS QANLU Data. Data augmentation improves the performance of QA models.}
\label{tbl:data_aug}
\end{table}
%%%%%%%%%%%%%%%%%%%%%%%%%
%%%%%%%%%%%%%%%%%%%%%%%%%
%%%%%%%%%%%%%%%%%%%%%%%%%

\bibliographystyle{IEEEbib}
%\bibliography{ref}

\begin{thebibliography}{10}

\bibitem{10.1007/978-3-030-01424-7_27}
Chuanqi Tan, Fuchun Sun, Tao Kong, Wenchang Zhang, Chao Yang, and Chunfang Liu,
\newblock ``A survey on deep transfer learning,''
\newblock in {\em Artificial Neural Networks and Machine Learning -- ICANN
  2018}, V{\v{e}}ra K{\r{u}}rkov{\'a}, Yannis Manolopoulos, Barbara Hammer,
  Lazaros Iliadis, and Ilias Maglogiannis, Eds., 2018, pp. 270--279.

\bibitem{zhuang2020comprehensive}
Fuzhen Zhuang, Zhiyuan Qi, Keyu Duan, Dongbo Xi, Yongchun Zhu, Hengshu Zhu, Hui
  Xiong, and Qing He,
\newblock ``A comprehensive survey on transfer learning,'' 2020.

\bibitem{Alyafeai2020ASO}
Zaid Alyafeai, Maged~S. Al-shaibani, and I.~Ahmad,
\newblock ``A survey on transfer learning in natural language processing,''
\newblock {\em ArXiv}, vol. abs/2007.04239, 2020.

\bibitem{pennington2014glove}
Jeffrey Pennington, Richard Socher, and Christopher~D. Manning,
\newblock ``Glove: Global vectors for word representation,''
\newblock in {\em Empirical Methods in Natural Language Processing (EMNLP)},
  2014, pp. 1532--1543.

\bibitem{NIPS2013_5021}
Tomas Mikolov, Ilya Sutskever, Kai Chen, Greg~S Corrado, and Jeff Dean,
\newblock ``Distributed representations of words and phrases and their
  compositionality,''
\newblock in {\em Advances in Neural Information Processing Systems 26},
  C.~J.~C. Burges, L.~Bottou, M.~Welling, Z.~Ghahramani, and K.~Q. Weinberger,
  Eds., pp. 3111--3119. 2013.

\bibitem{devlin-etal-2019-bert}
Jacob Devlin, Ming-Wei Chang, Kenton Lee, and Kristina Toutanova,
\newblock ``{BERT}: Pre-training of deep bidirectional transformers for
  language understanding,''
\newblock in {\em Proceedings of the 2019 Conference of the North {A}merican
  Chapter of the Association for Computational Linguistics: Human Language
  Technologies, Volume 1}, June 2019, pp. 4171--4186.

\bibitem{Raffel2019ExploringTL}
Colin Raffel, Noam Shazeer, Adam Roberts, Katherine Lee, Sharan Narang, Michael
  Matena, Yanqi Zhou, W.~Li, and P.~Liu,
\newblock ``Exploring the limits of transfer learning with a unified
  text-to-text transformer,''
\newblock {\em ArXiv}, vol. abs/1910.10683, 2019.

\bibitem{DBLP:journals/corr/abs-1806-08730}
Bryan McCann, Nitish~Shirish Keskar, Caiming Xiong, and Richard Socher,
\newblock ``The natural language decathlon: Multitask learning as question
  answering,''
\newblock {\em CoRR}, vol. 1806.08730, 2018.

\bibitem{DBLP:journals/corr/abs-1902-10909}
Qian Chen, Zhu Zhuo, and Wen Wang,
\newblock ``{BERT} for joint intent classification and slot filling,''
\newblock {\em CoRR}, vol. 1902.10909, 2019.

\bibitem{krone-etal-2020-learning}
Jason Krone, Yi~Zhang, and Mona Diab,
\newblock ``Learning to classify intents and slot labels given a handful of
  examples,''
\newblock in {\em Proceedings of the 2nd Workshop on Natural Language
  Processing for Conversational AI}, 2020, pp. 96--108.

\bibitem{coope2020spanconvert}
Sam Coope, Tyler Farghly, Daniela Gerz, Ivan Vulić, and Matthew Henderson,
\newblock ``Span-convert: Few-shot span extraction for dialog with pretrained
  conversational representations,'' 2020.

\bibitem{gao2019dialog}
Shuyang Gao, Abhishek Sethi, Sanchit Agarwal, Tagyoung Chung, and Dilek
  Hakkani-Tur,
\newblock ``Dialog state tracking: A neural reading comprehension approach,''
\newblock {\em arXiv preprint arXiv:1908.01946}, 2019.

\bibitem{gao2020machine}
Shuyang Gao, Sanchit Agarwal, Tagyoung Chung, Di~Jin, and Dilek Hakkani-Tur,
\newblock ``From machine reading comprehension to dialogue state tracking:
  Bridging the gap,''
\newblock {\em arXiv preprint arXiv:2004.05827}, 2020.

\bibitem{zhou2019multi}
Li~Zhou and Kevin Small,
\newblock ``Multi-domain dialogue state tracking as dynamic knowledge graph
  enhanced question answering,''
\newblock {\em arXiv preprint arXiv:1911.06192}, 2019.

\bibitem{bapna2017towards}
Ankur Bapna, Gokhan Tur, Dilek Hakkani-Tur, and Larry Heck,
\newblock ``Towards zero-shot frame semantic parsing for domain scaling,''
\newblock {\em arXiv preprint arXiv:1707.02363}, 2017.

\bibitem{bhathiya2020meta}
Hemanthage~S Bhathiya and Uthayasanker Thayasivam,
\newblock ``Meta learning for few-shot joint intent detection and
  slot-filling,''
\newblock in {\em Proceedings of the 2020 5th International Conference on
  Machine Learning Technologies}, 2020, pp. 86--92.

\bibitem{shah2019robust}
Darsh~J Shah, Raghav Gupta, Amir~A Fayazi, and Dilek Hakkani-Tur,
\newblock ``Robust zero-shot cross-domain slot filling with example values,''
  2019.

\bibitem{madotto2020language}
Andrea Madotto,
\newblock ``Language models as few-shot learner for task-oriented dialogue
  systems,''
\newblock {\em arXiv preprint arXiv:2008.06239}, 2020.

\bibitem{NIPS2017_7181}
Ashish Vaswani, Noam Shazeer, Niki Parmar, Jakob Uszkoreit, Llion Jones,
  Aidan~N Gomez, \L~ukasz Kaiser, and Illia Polosukhin,
\newblock ``Attention is all you need,''
\newblock in {\em Advances in Neural Information Processing Systems 30},
  I.~Guyon, U.~V. Luxburg, S.~Bengio, H.~Wallach, R.~Fergus, S.~Vishwanathan,
  and R.~Garnett, Eds., pp. 5998--6008. 2017.

\bibitem{rajpurkar-etal-2018-know}
Pranav Rajpurkar, Robin Jia, and Percy Liang,
\newblock ``Know what you don{'}t know: Unanswerable questions for {SQ}u{AD},''
\newblock in {\em Proceedings of the 56th Annual Meeting of the Association for
  Computational Linguistics}, pp. 784--789.

\bibitem{hemphill-etal-1990-atis}
Charles~T. Hemphill, John~J. Godfrey, and George~R. Doddington,
\newblock ``The {ATIS} spoken language systems pilot corpus,''
\newblock in {\em Speech and Natural Language: Proceedings of a Workshop Held
  at Hidden Valley, {P}ennsylvania, June 24-27,1990}, 1990.

\bibitem{5700816}
G.~{Tur}, D.~{Hakkani-Tür}, and L.~{Heck},
\newblock ``What is left to be understood in atis?,''
\newblock in {\em 2010 IEEE Spoken Language Technology Workshop}, 2010, pp.
  19--24.

\bibitem{Wolf2019HuggingFacesTS}
Thomas Wolf, Lysandre Debut, Victor Sanh, Julien Chaumond, Clement Delangue,
  Anthony Moi, Pierric Cistac, Tim Rault, Rémi Louf, Morgan Funtowicz, Joe
  Davison, Sam Shleifer, Patrick von Platen, Clara Ma, Yacine Jernite, Julien
  Plu, Canwen Xu, Teven~Le Scao, Sylvain Gugger, Mariama Drame, Quentin Lhoest,
  and Alexander~M. Rush,
\newblock ``Huggingface's transformers: State-of-the-art natural language
  processing,''
\newblock {\em ArXiv}, vol. abs/1910.03771, 2019.

\bibitem{sanh2020distilbert}
Victor Sanh, Lysandre Debut, Julien Chaumond, and Thomas Wolf,
\newblock ``Distilbert, a distilled version of bert: smaller, faster, cheaper
  and lighter,'' 2020.

\bibitem{Lan2020ALBERTAL}
Zhenzhong Lan, Mingda Chen, Sebastian Goodman, Kevin Gimpel, Piyush Sharma, and
  Radu Soricut,
\newblock ``Albert: A lite bert for self-supervised learning of language
  representations,''
\newblock {\em ArXiv}, vol. abs/1909.11942, 2020.

\bibitem{zhang2019joint}
Chenwei Zhang, Yaliang Li, Nan Du, Wei Fan, and Philip~S Yu,
\newblock ``Joint slot filling and intent detection via capsule neural
  networks,''
\newblock in {\em Proceedings of the 57th Annual Meeting of the Association for
  Computational Linguistics (ACL)}, 2019.

\end{thebibliography}

\end{document}